\pdfoutput=1
\documentclass[acmsmall,screen,nonacm]{acmart}

\setcopyright{none}
\renewcommand\footnotetextcopyrightpermission[1]{}

\usepackage{booktabs}
\usepackage{xcolor}

\begin{document}

\title{Commit or Abstain? System One Decision Models, LLMs, and Missing Legal Authority}

\author{M. Mikail Demir}
\affiliation{
  \institution{Department of Information Science, University at Albany, SUNY}
  \city{Albany}
  \state{NY}
  \country{USA}}
\email{mdemir@albany.edu}

\begin{abstract}
Given the enormous number of claims to be checked, verifiable legal AI requires scalable, cost-effective models that answer the practical needs of law. System One decision models are built on large language models but, instead of generating text, return a calibrated probability for each option. This suits the verification of legal authority, where the question is closed: a given holding, or none of them. We test whether such models abstain when the correct authority is missing. We extend CaseHOLD with an abstention condition in which the correct holding is replaced in 1,670 of 5,314 items, and compare four decision models with six generative LLMs. When no option is correct, generative LLMs still pick one in 38--86\% of items; the most cautious decision model does so in 30\%. On standard CaseHOLD, a decision model matches the best generative LLM (88.1\% vs.\ 88.2\%) at 1/55 of the cost. Accepting that LLM's answer only when a decision model agrees halves its wrong citations, and distilling a decision model's probabilities into Qwen3-8B, without labels, yields a well-calibrated local model. Legal AI should be judged by its wrong citations, not by accuracy alone.
\end{abstract}


\maketitle

\begin{flushright}
\begin{minipage}{0.75\linewidth}
\itshape ``Resolve to be honest at all events; and if in your own judgment you cannot be an honest lawyer, resolve to be honest without being a lawyer.''\\[2pt]
\upshape\hfill --- Abraham Lincoln, \emph{Notes for a Law Lecture} (c.~1850)
\end{minipage}
\end{flushright}

\section{Introduction}
\label{sec:intro}

\begin{figure}[t]
  \centering
  \begin{minipage}[c]{0.43\linewidth}
    \scriptsize
    \fbox{\parbox{0.95\linewidth}{
      \textbf{(a) The correct holding is removed}\\[3pt]
      \textit{``\ldots It is not the province of this Court, at this stage, to introduce a remedy that in hindsight appears better for the appellant as a matter of equity. See, e.g., In re Bernal, 207 F.3d 595 (9th Cir.~2000) (\textless HOLDING\textgreater)''}\\[3pt]
      A.\ \ldots the fourth amendment remedy sought is suppression\\
      B.\ \ldots the remedy in a prison conditions case must remedy actual injuries\ldots\\
      C.\ \ldots no legitimate basis for distinction may be found between the situation where the claim is not timely raised and where the remedy sought was not timely pursued\\
      D.\ \ldots retroactive award of benefits proper remedy\ldots\\
      E.\ \ldots rendition is remedy for no evidence\\
      F.\ None of these \textbf{(correct)}\\[3pt]
      \textcolor[HTML]{EB6834}{\textbf{All six LLMs:}} C\\
      \textcolor[HTML]{2A78D6}{\textbf{Three of four System One models:}} F
    }}
  \end{minipage}\hfill
  \begin{minipage}[c]{0.55\linewidth}
    \centering
    \textbf{\scriptsize (b) How often models still pick a holding}\\[2pt]
    \includegraphics[width=\linewidth]{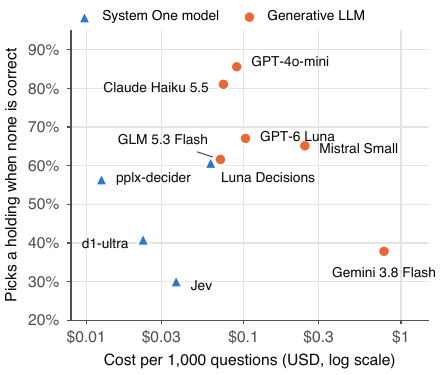}
  \end{minipage}
  \caption{When the correct authority is missing, most models still cite one. (a) A CaseHOLD item whose true holding was replaced by a similar holding from another case (C); every generative LLM chose it. (b) Share of the 1,670 such items in which each model picked a holding instead of abstaining, against cost per 1,000 questions. System One models are the four cheapest and include the most cautious; among generative LLMs, more spending does not reliably buy caution.}
  \label{fig:teaser}
\end{figure}

Language models are now used across legal practice, and most visibly in legal research. Law has also become one of the best-documented domains of their failure. Since late 2022, courts have sanctioned lawyers and pro se litigants for filings that cite authorities that do not exist or do not support the propositions they are cited for~\cite{mata2023avianca,billups2025}. Each such filing adds work to an already overburdened court system. Someone has to check the citations, and courts must then hold hearings and decide on sanctions. The problem now reaches the bench, with judges' own orders among the documented cases~\cite{charlotin2026}.

The problem is not limited to general-purpose chatbots. In a preregistered evaluation, commercial legal research tools built on retrieval-augmented generation, some marketed as hallucination-free, still hallucinated on 17\% to 33\% of queries~\cite{magesh2025hallucinationfree}; general-purpose models did so at least 58\% of the time~\cite{dahl2024fictions}. Courts now see the same failure in practice. In October 2026, a federal court reviewed filings whose author admitted taking propositions from a Westlaw AI tool without confirming them~\cite{mercante2026}. Nor is the problem fading with newer models. Across eight generations of ChatGPT models, citation hallucination rates did not consistently decline, and GPT-5.1 produced hallucinated citations more often than the best mid-2024 GPT-4o release~\cite{liu2026citations}.

Institutions are responding. California requires its courts to adopt generative-AI use policies that address hallucinated output~\cite{calrule10430}, and the State Bar of California has proposed amending its rule on candor to the tribunal so that lawyers must verify that cited authorities exist and are accurate~\cite{calbar2026amendments}. Texas has proposed that signers of filings represent that their citations, quotations, and holdings are not fabricated~\cite{texas2026misc}. Appellate courts now sanction unverified citations with fees, fines, and referrals to the bar~\cite{noland2025,whiting2026}. Yet the cases keep growing, and one public tracker lists 2,203 court matters involving AI hallucinations, 1,264 of them from 2026 alone~\cite{charlotin2026}. Fabrication is only part of the problem. In 933 of these matters, a real authority was cited for a proposition it does not support~\cite{charlotin2026}.

Technical responses mostly add more language-model machinery: retrieval augmentation, better training data, model editing, decoding strategies~\cite{huang2025survey}, and agentic verification.
These efforts work against a structural pressure. \citet{kalai2025hallucinate} trace hallucinations to a simple classification problem, deciding whether a statement is valid. Models make errors on this problem, and the errors persist because training and evaluation reward guessing. Under accuracy-style scoring, a model that always guesses scores higher than one that admits it is unsure. In legal writing, the guess is the costly outcome, a confident citation to the wrong authority. Legal studies observe the same tendency. When told to stop if no factual basis existed, most models produced arguments anyway~\cite{zhang2025abstention}.

Most evaluations ask whether a model picks the right authority from a set of candidates, but in practice the right authority is often not among them. A useful system should then say so rather than choose the closest option. We therefore ask whether a model, when the correct authority is missing, commits to a wrong one or abstains. The two errors are not equally costly. A wrong commitment can reach a filing; an unnecessary abstention costs a lawyer an extra lookup. Accuracy treats them the same. What matters instead is whether a model's confidence can tell a user which answers to check.

Legal NLP posed many of these problems as classification well before large language models. CaseHOLD asks which of five holdings a court gave for a cited case~\cite{zheng2021casehold}; ContractNLI asks whether a contract entails, contradicts, or does not mention a hypothesis~\cite{koreeda2021contractnli}; MAUD poses expert questions about merger agreements~\cite{wang2023maud}. These tasks were addressed with fine-tuned encoders. Since large language models arrived, the same tasks are mostly posed to general-purpose generative models.

The underlying need has not changed. Lawyers and courts need to check, at scale, whether a specific authority supports a specific proposition, and to be told when it does not. This calls for precision and calibrated confidence more than open-ended reasoning. A verification model should decide only when the evidence is sufficient~\cite{atanasova2022insufficient}, and small models trained for a narrow verification task can match GPT-4 at a fraction of its cost~\cite{tang2024minicheck}. If more reasoning alone solved the problem, the strongest models would not continue to hallucinate~\cite{liu2026citations}. In legal case-based reasoning, reasoning models even spend more computation on wrong answers than on right ones~\cite{zhang2026thinking}.

System One decision models revisit the classification framing with modern models. TypeSafe released the first of them, Jev, in September 2026~\cite{typesafe2026introducing}. The name follows the fast, intuitive System~1 of \citet{kahneman2011thinking}, and commentators soon began calling them decision models~\cite{willison2026jev}. They are themselves built on large language models, but instead of generating text they return a typed decision and a probability for each option. By late September, an analysis of GitHub found 2,170 public projects using Jev~\cite{ling2026wild}, and a community index now tracks more than 170 decision models from many developers~\cite{decisionindex2026}. This design fits the task of checking an authority well. The question is closed (this holding, or none of them), so there is nothing to write, and the probability states how sure the model is, which is what a user needs in order to decide what to check. TypeSafe trains its Jev model with reinforcement learning for calibrated decisions (RLCD), which optimizes these probabilities against outcomes rather than human preferences~\cite{typesafe2026systemone}. Whether such models are more trustworthy than generative LLMs where it matters most in law, when the right authority is not available, has not been tested.

We examine this with CaseHOLD. We add a condition in which the correct holding is sometimes replaced by a similar holding from another case, and a sixth option states that none of the holdings applies. We compare four System One decision models with six generative LLMs on all 5,314 test items. We ask how often each model commits to a holding when none is correct, what its caution costs, and whether the two kinds of model can cover for each other.

Our contributions are:
\begin{itemize}
  \item \textbf{A test of abstention for legal authority.} We extend CaseHOLD so that in 1,670 of its 5,314 test items the correct holding is replaced by a similar holding from another case, and measure how often a model cites a holding when none applies (Figure~\ref{fig:teaser}a).
  \item \textbf{A comparison across model types and prices.} We evaluate four System One decision models and six generative LLMs from five vendors, reporting accuracy, calibration, abstention, and cost (Figure~\ref{fig:teaser}b).
  \item \textbf{A cost-aware view of combining them.} We show how pairing a System One model with a generative LLM trades wrong citations against items sent for human review, and frame the choice of operating point by the unequal cost of the two errors.
  \item \textbf{Local deployment and checks on what the results mean.} Trained only on a System One model's probabilities, with no labeled data, a local Qwen3-8B model learns to answer and abstain like its teacher and becomes well calibrated. Contamination tests extending \citet{arvin2025holdings} show that models know much of the case law but do not depend on it.
\end{itemize}

Code, data, and the outputs of every model are available at \url{https://github.com/mmikaildemir/commit-or-abstain}.

\section{Related Work}
\label{sec:related}

\subsection{Legal Datasets with Verifiable Answers}

Legal NLP has a long line of datasets whose answers can be checked against a source. Many of them come from contracts. ContractNLI asks whether a contract entails, contradicts, or says nothing about a hypothesis~\cite{koreeda2021contractnli}, CUAD marks the clauses lawyers look for in contract review~\cite{hendrycks2021cuad}, and MAUD asks expert questions about merger agreements~\cite{wang2023maud}. COLIEE has run shared tasks on retrieving and entailing statutes and case law for years~\cite{goebel2024coliee}. In CaseHOLD, the dataset we use, a model picks which of five holdings a court gave for a cited case~\cite{zheng2021casehold}; Section~\ref{sec:method} describes it in detail. LexGLUE~\cite{chalkidis2022lexglue} and LegalBench~\cite{guha2023legalbench} later gathered tasks like these into suites for evaluating language models, and LawBench did the same for Chinese law~\cite{fei2024lawbench}. In almost all of them, the correct answer is among the candidates and every error counts the same. Neither is true when a lawyer checks a citation.

\subsection{Abstention and Unanswerable Questions}

Outside law, SQuAD~2.0 added questions that look answerable but are not, so that a system had to learn when to stay silent~\cite{rajpurkar2018squad2}. Later work asked the same of language models with multiple-choice questions. Models lose much of their accuracy when \textit{none of the above} is the right option~\cite{tam2025nota}, and they keep choosing an option even when every option is wrong~\cite{goral2025option}. Across twenty datasets, AbstentionBench even finds that reasoning fine-tuning can make abstention worse~\cite{kirichenko2025abstentionbench}. \citet{kalai2025hallucinate} argue that this follows from how models are graded, since a guess is rewarded over an admission of uncertainty. Our abstain condition brings this test to legal authority, where a confident guess becomes a citation to the wrong holding.

\subsection{System One Decision Models}

System One models answer typed questions with a probability for each option instead of generated text~\cite{typesafe2026systemone}. Jev, the first of them, is trained with RLCD, and its architecture and training remain proprietary. Adoption has been fast. An analysis of 2,170 public projects finds Jev used mainly for judging attributes and scoring, as a component inside larger workflows~\cite{ling2026wild}. A community index now tracks more than 170 decision models from many developers~\cite{decisionindex2026}.

Open work has taken two routes to the same behavior. One trains for it directly, implementing Jev's objective with strictly proper scoring rules~\cite{gao2026openjevrlcd}. The other reads it out of existing language models, whose next-token probabilities over the option letters already behave as a decision model, and light fine-tuning closes most of the remaining gap~\cite{geuter2026innerjev,li2026llmasjev}. A third recipe distills a model's own post-reasoning answer into a single-pass readout~\cite{duan2026general}. Open-weight models such as Laya, a 421M-parameter encoder, now exist alongside Jev~\cite{nandakishore2026laya}.

Independent evaluations have appeared only in recent weeks. Jev is competitive with frontier generative LLMs on knowledge and commonsense benchmarks but falls behind on multi-step arithmetic~\cite{li2026specialized,deusser2026jev}. Decision models do best when the answer can be resolved from the evidence given, but often overstate the probability of their most likely answer~\cite{duan2026general}, and their calibration shifts with the wording of a question~\cite{sharma2026typedbench}. Two studies bear directly on abstention. In medical question answering, most evaluated models struggle to recognize when the correct answer is missing~\cite{du2026medicine}. Jev's confidence can also stay high when no answer-relevant information is given at all~\cite{shankaranarayana2026selfknowledge}.

Decision models are also combined with other models. They serve as gates in front of generative LLMs~\cite{li2026fastmodels}, and escalating their uncertain cases to a generative LLM can match the larger model's accuracy at close to the decision model's cost~\cite{rafe2026gates}, in the line of earlier cascades and routers~\cite{chen2023frugalgpt,ong2025routellm}. Jev's judgments have also been distilled into small task-specific evaluators for chess and passage reranking~\cite{sadoun2026probabilities}, though not, to our knowledge, into a general-purpose generative LLM.

In law, \citet{zhang2026samescores} compare Jev with nine generative LLMs on ContractNLI and find it the cheapest and fastest, though less accurate than hosted models. No prior work evaluates decision models on choosing legal authority, or combines them with generative LLMs to limit wrong citations.

\section{Method}
\label{sec:method}

\begin{figure}[t]
  \centering
  \includegraphics[width=\linewidth]{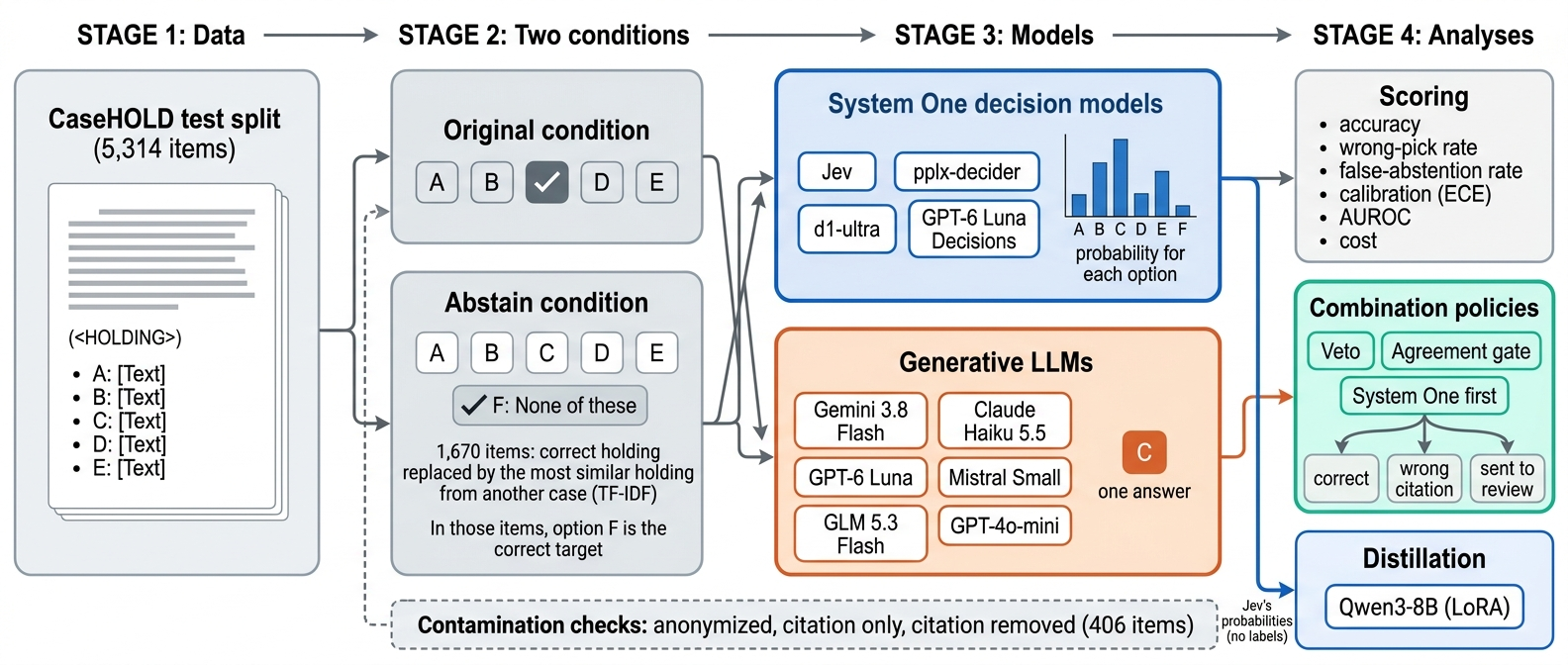}
  \caption{Overview of the study. CaseHOLD test items are presented in two conditions. In the abstain condition, the correct holding is replaced in 1,670 of 5,314 items, and a sixth option, none of these, becomes correct. System One decision models return a probability for each option, and generative LLMs return one answer. Their outputs are scored, combined through three policies, and used to train a local model.}
  \label{fig:method}
\end{figure}

Figure~\ref{fig:method} gives an overview of the study.

\subsection{Task and Data}

CaseHOLD is built from U.S. court opinions in the Harvard Law Library case law corpus~\cite{zheng2021casehold}. Courts often follow a citation with a parenthetical stating the cited case's holding, for example \textit{(holding that consideration exists where a customer must visit the store)}. CaseHOLD replaces this parenthetical with the token \textless HOLDING\textgreater. Each item consists of a short excerpt from the citing opinion and five candidate holdings. One is the holding the court wrote, and the other four are holdings from other citations with the most similar wording. We use the full test split of 5,314 items.

\subsection{Two Conditions}

The \emph{original} condition is CaseHOLD as published, with five options and one correct answer.

The \emph{abstain} condition adds a sixth option, \textit{None of the listed holdings is the one the cited case is being cited for in this passage}. In a random 31\% of items (1,670), the correct holding is replaced by another real holding, and the sixth option becomes the correct answer. The replacement is selected with CaseHOLD's procedure for wrong options. It is the holding from another test item with the highest TF-IDF cosine similarity to the removed holding, subject to CaseHOLD's similarity cap of 0.75. Replacements have a mean similarity of 0.28 to the removed holding, compared with 0.30 for CaseHOLD's own wrong options. In the remaining 3,644 items the correct holding is kept, and the sixth option is wrong.

\subsection{Models}

We compare two kinds of model, both built on large language models, that differ in their output (Table~\ref{tab:models}). \emph{System One decision models} receive the passage as state and a single choice question whose options are the holdings, and they return a probability for each option. We evaluate Jev~\cite{typesafe2026systemone}, Perplexity's pplx-decider, Liquid AI's d1-ultra, and OpenAI's GPT-6 Luna Decisions through OpenRouter's decisions interface. \emph{Generative LLMs} receive the same instruction, passage, and options as a chat message and return the key of one option. The six generative LLMs come from five vendors and cover a range of prices. They include GPT-4o-mini, which \citet{arvin2025holdings} evaluated on CaseHOLD, and GLM 5.3 Flash, whose price per question is close to Jev's.

\begin{table}[t]
  \centering
  \small
  \caption{Models evaluated. Cost is the amount billed per 1,000 questions in the original condition; batch-processed models were billed at batch rates.}
  \label{tab:models}
  \begin{tabular}{llllr}
    \toprule
    Model & Type & Provider & Mode & \$ / 1,000 \\
    \midrule
    Jev 1.14 & System One & TypeSafe & sync & 0.035 \\
    pplx-decider v1.2 & System One & Perplexity & sync & 0.012 \\
    d1-ultra & System One & Liquid AI & sync & 0.021 \\
    GPT-6 Luna Decisions & System One & OpenAI & sync & 0.058 \\
    \midrule
    Gemini 3.8 Flash & Generative LLM & Google & batch & 0.659 \\
    Claude Haiku 5.5 & Generative LLM & Anthropic & batch & 0.066 \\
    GPT-6 Luna & Generative LLM & OpenAI & batch & 0.077 \\
    Mistral Small 2603 & Generative LLM & Mistral AI & batch & 0.234 \\
    GLM 5.3 Flash & Generative LLM & Z.ai & sync & 0.065 \\
    GPT-4o-mini & Generative LLM & OpenAI & sync & 0.084 \\
    \bottomrule
  \end{tabular}
\end{table}

\subsection{Prompts and Settings}

All models receive the same one-sentence instruction, which states that a court opinion cites a case, that the parenthetical describing its holding has been replaced by \textless HOLDING\textgreater, and that the task is to identify the holding the citing court gave. Decision models receive the passage as state and the holdings as the options of a choice question. Generative LLMs receive the passage and lettered options in one message and reply with a JSON object containing only the option key. This format is enforced with structured output, since some models otherwise produced an explanation first. Generative LLMs are not asked for a confidence score; Appendix~\ref{app:pilot} reports a pilot with stated confidence.

Generative LLMs run at temperature 0 with the lowest reasoning setting each model offers above none; GPT-4o-mini has no reasoning setting. We record reasoning tokens and the billed cost of every call.\footnote{One passage, from a criminal case, was refused by the content filter of Claude Haiku 5.5's provider, so that model's abstain-condition results cover 5,313 items.} The full prompts are in Appendix~\ref{app:prompts}.

\subsection{Metrics}

We report accuracy and macro F1 in both conditions. In the abstain condition, two error rates are reported separately. The \emph{wrong-pick rate} is the share of items without a correct holding in which the model picked a holding. The \emph{false-abstention rate} is the share of items with a correct holding in which the model chose the none option. For decision models, we also report expected calibration error over ten bins and the AUROC of the none probability for separating items with and without a correct holding. Confidence intervals are 95\% percentile bootstrap intervals with 2,000 resamples. Paired comparisons between models use exact McNemar tests on the same items.

\subsection{Combining Decision Models and Generative LLMs}

We evaluate three policies that combine a decision model with a generative LLM, all on the abstain condition. In the \emph{veto} policy, the generative LLM's answer is accepted unless the decision model's probability of none exceeds a threshold. In the \emph{agreement gate}, an answer is accepted only when both models choose the same option, and other items are sent for review. This follows the observation that hard items tend to defeat some models but not others~\cite{demir2026parcel}. In \emph{System One first}, the decision model answers when its top probability exceeds a threshold, and the generative LLM is called only for the remaining items, which reduces cost.

Each item ends as a correct answer, a wrong citation, or a review, meaning an item flagged for a person although the correct holding was present. We score a policy by $k \cdot \text{wrong citations} + \text{reviews}$, where $k$ is the cost of a wrong citation relative to an unnecessary review. We know of no empirical estimate of $k$ for legal research. The values $k = 1, 5, 10$ are therefore illustrative, and Figure~\ref{fig:tradeoff} shows the full trade-off. Thresholds are tuned on a random 30\% of items, and results are reported on the remaining 70\% (3,720 items).

\subsection{Contamination Checks}

Since CaseHOLD is built from public opinions, a model may recognize a cited case instead of reasoning from the passage. Following \citet{arvin2025holdings}, we test this with three variants of each passage, one with case names and citations anonymized, one showing only the cited case's citation, and one with that citation removed. Appendix~\ref{app:contamination} describes the procedure and results.

\subsection{Distillation}

We test whether a decision model's probabilities can train a model that runs locally. Qwen3-8B is fine-tuned with LoRA on 10,000 items from CaseHOLD's training split in the abstain condition. We compare three versions: the base model, a model trained only to match Jev's probabilities, and a model trained on gold labels. The first two use no labeled data. Each model's answer distribution is read from its next-token probabilities over the option letters~\cite{geuter2026innerjev,li2026llmasjev}, so that its calibration can be measured in the same way as a decision model's. Earlier work distilled Jev into task-specific evaluators~\cite{sadoun2026probabilities}; here the student is a general-purpose generative LLM. Each version was trained once on a single NVIDIA RTX A6000 (48\,GB). A training run took about 90 minutes, and evaluation on the full test split about 13 minutes. Training settings are in Appendix~\ref{app:distill}.

\section{Results}
\label{sec:results}

\begin{table}[t]
  \centering
  \small
  \caption{Results on all 5,314 CaseHOLD test items. \emph{Wrong pick}: share of the 1,670 abstain-condition items without a correct holding in which the model picked a holding (95\% bootstrap CI). \emph{False abstention}: share of the 3,644 items with a correct holding in which the model abstained. AUROC of the probability of none is available only for models that return probabilities. 95\% CIs for accuracy are within $\pm$1.4 points.}
  \label{tab:main}
  \begin{tabular}{lrrrrrr}
    \toprule
    & \multicolumn{2}{c}{Accuracy (\%)} & & & & \\
    \cmidrule(lr){2-3}
    Model & Original & Abstain & Wrong pick (\%) & False abst.\ (\%) & AUROC & \$ / 1,000 \\
    \midrule
    \multicolumn{7}{l}{\textit{System One decision models}} \\
    Jev 1.14 & 78.3 & 63.9 & \textbf{30.1} [27.8, 32.3] & 26.2 & 0.784 & 0.035 \\
    pplx-decider v1.2 & \textbf{88.1} & 73.1 & 56.5 [54.2, 58.9] & 1.8 & \textbf{0.887} & \textbf{0.012} \\
    d1-ultra & 84.9 & 73.6 & 41.0 [38.6, 43.3] & 9.5 & 0.856 & 0.021 \\
    GPT-6 Luna Decisions & 72.2 & 58.2 & 60.8 [58.4, 63.0] & 9.5 & 0.726 & 0.058 \\
    \midrule
    \multicolumn{7}{l}{\textit{Generative LLMs}} \\
    Gemini 3.8 Flash & \textbf{88.2} & \textbf{78.4} & 37.8 [35.5, 40.1] & 4.2 & -- & 0.659 \\
    Claude Haiku 5.5 & 75.2 & 56.5 & 81.1 [79.1, 82.9] & 2.6 & -- & 0.066 \\
    GPT-6 Luna & 77.6 & 62.1 & 67.1 [64.7, 69.5] & 2.8 & -- & 0.077 \\
    Mistral Small 2603 & 70.9 & 57.0 & 65.1 [62.8, 67.4] & 7.7 & -- & 0.234 \\
    GLM 5.3 Flash & 74.6 & 60.8 & 61.6 [59.2, 64.0] & 7.0 & -- & 0.065 \\
    GPT-4o-mini & 67.4 & 50.1 & 85.6 [83.9, 87.3] & 2.1 & -- & 0.084 \\
    \bottomrule
  \end{tabular}
\end{table}

\subsection{Accuracy and Cost}

On the original task, the best decision model and the best generative LLM are tied. pplx-decider reaches 88.1\% and Gemini 3.8 Flash 88.2\% (difference $-0.2$ points, 95\% CI $[-1.2, 0.8]$, $p = 0.74$), at about 1/55 of the cost (Table~\ref{tab:main}). Three of the four decision models are more accurate than every generative LLM except Gemini. Jev, the decision model whose price is closest to GLM 5.3 Flash, is 3.7 points more accurate ($[2.6, 4.8]$, $p < 10^{-10}$). Macro F1 tracks accuracy within 0.002 for every model. All models lose accuracy in the abstain condition, by 9.9 to 18.7 points.

\subsection{Committing When No Holding Is Correct}

When the correct holding is missing, most models still pick one (Figure~\ref{fig:teaser}). Among generative LLMs, the wrong-pick rate ranges from 37.8\% for Gemini 3.8 Flash to 85.6\% for GPT-4o-mini, the model \citet{arvin2025holdings} evaluated on CaseHOLD. Among decision models, it ranges from 30.1\% for Jev to 60.8\% for GPT-6 Luna Decisions. Jev picks a wrong holding 7.7 points less often than Gemini ($[-10.0, -5.3]$, $p < 10^{-9}$) and 31.5 points less often than GLM 5.3 Flash, the generative LLM closest to its price ($[-34.0, -29.1]$, $p < 10^{-100}$).

Caution has a price. Jev abstains on 26.2\% of items that do have a correct holding, while every generative LLM does so on fewer than 8\%. The probabilities of the decision models separate the two kinds of item well (AUROC 0.73 to 0.89), so a user can set a threshold on them.

\begin{figure}[t]
  \centering
  \includegraphics[width=0.85\linewidth]{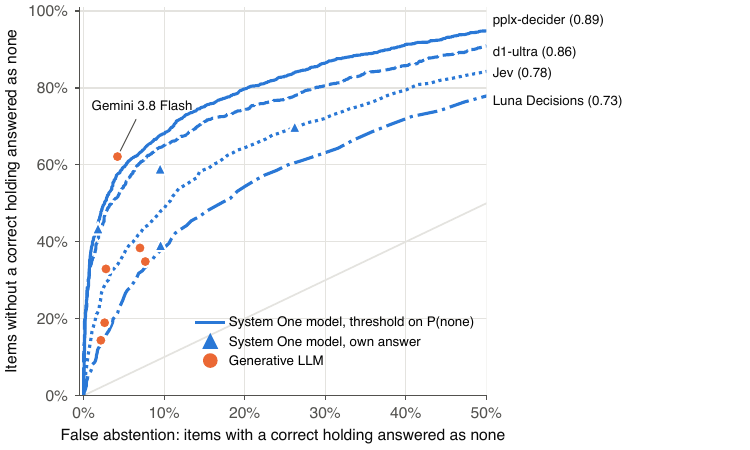}
  \caption{Abstention as a threshold on the probability of none, on all 5,314 abstain-condition items. Each curve traces one System One model as the threshold moves; AUROC in parentheses. Triangles mark each decision model's own answer, and circles mark the generative LLMs, which offer a single operating point. The x-axis is cut at 50\%.}
  \label{fig:roc}
\end{figure}

Figure~\ref{fig:roc} shows the choices such a threshold offers. A generative LLM gives a single operating point, while a decision model can be moved along its curve. At the false-abstention rate of every generative LLM except Gemini 3.8 Flash, pplx-decider and d1-ultra both recognize more of the items without a correct holding. Gemini lies above all four curves. At its false-abstention rate of 4.2\%, pplx-decider recognizes 57.7\% of these items, against 62.2\% for Gemini.

\subsection{Combining Decision Models and Generative LLMs}

\begin{figure}[t]
  \centering
  \includegraphics[width=0.92\linewidth]{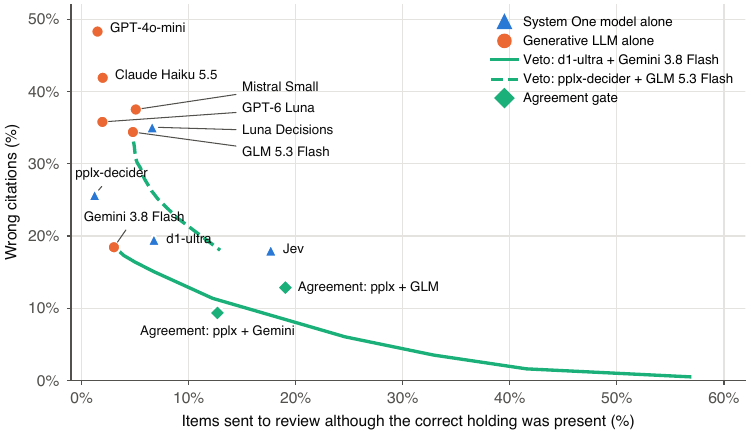}
  \caption{Wrong citations against unnecessary reviews on the abstain condition (held-out 70\% of items, $n = 3{,}720$). Each point is a model alone. The curves show the veto policy as its threshold varies; diamonds show the agreement gate. Moving down a curve buys fewer wrong citations with more review.}
  \label{fig:tradeoff}
\end{figure}

Combining the two kinds of model lowers wrong citations, but sends more items to a person for review (Figure~\ref{fig:tradeoff}). The veto and System One first policies need a threshold, which we chose on a random 30\% of items (Section~\ref{sec:method}). To keep that choice from flattering the results, every number in this section comes from the remaining 70\% (3,720 items), which played no part in choosing it.

Take Gemini 3.8 Flash as an example. Alone, it cites a wrong holding on 18.4\% of items. With the agreement gate, we accept Gemini's answer only when pplx-decider gives the same answer. This cuts wrong citations by half, to 9.4\%. The price is review: 12.7\% of items now go to a person even though the correct holding was there, against 3.0\% for Gemini alone. The share of correct answers barely changes (78.0\% against 78.5\%), and the extra cost is about 1\%.

Cheaper pairs also help. Gating GLM 5.3 Flash with pplx-decider gives fewer wrong citations than Gemini alone (12.8\%), at about one-ninth of Gemini's cost.

Which policy is best depends on how much worse a wrong citation is than an unnecessary review. If the two count the same ($k = 1$), a decision model alone is nearly as good as anything else. d1-ultra makes about as many wrong citations as Gemini (19.5\%) at 1/30 of the cost. If a wrong citation counts five times as much ($k = 5$), a combined policy is best for every pair we tested. At $k = 10$, the best policy for most pairs sends more than half of the items to review, and wrong citations fall to about 2\% or less.

\subsection{Contamination}

Models know much of the case law but do not depend on it. The strongest models pick the right holding from the citation alone well above chance, yet anonymizing case names and citations costs only 0.5 to 5.7 points. Appendix~\ref{app:contamination} gives the full results.

\subsection{Distillation}

\begin{table}[t]
  \centering
  \small
  \caption{Qwen3-8B on the abstain condition (all 5,314 test items) before fine-tuning, after fine-tuning on Jev's probabilities alone, and after fine-tuning on gold labels. Each version was trained once on 10,000 training items.}
  \label{tab:distill}
  \begin{tabular}{lrrrrr}
    \toprule
    Model & Accuracy (\%) & Wrong pick (\%) & False abst.\ (\%) & AUROC & ECE \\
    \midrule
    Qwen3-8B, no fine-tuning & 43.6 & 21.8 & 58.1 & 0.658 & 0.504 \\
    Qwen3-8B, Jev's probabilities & 63.8 & 31.2 & 24.6 & 0.777 & 0.025 \\
    Qwen3-8B, gold labels & 74.9 & 38.1 & 7.2 & 0.889 & 0.023 \\
    \midrule
    Jev 1.14 (teacher) & 63.9 & 30.1 & 26.2 & 0.784 & 0.056 \\
    \bottomrule
  \end{tabular}
\end{table}

This experiment asks whether a decision model's probabilities can be used to train a model that a firm or court could run on its own hardware. We fine-tuned Qwen3-8B in two ways, once to match Jev's probabilities and once on the gold labels, and compare both with the model before fine-tuning (Table~\ref{tab:distill}). The first version never sees a gold label. All three are evaluated on the full abstain condition.

Before fine-tuning, Qwen3-8B is not usable for this task. It answers none on most items, whether or not a correct holding is present, and its probabilities are far from calibrated (ECE 0.504).

Trained only on Jev's probabilities, Qwen3-8B behaves much like Jev. Its accuracy is 63.8\%, against Jev's 63.9\%. It picks a wrong holding and abstains wrongly about as often as Jev, and its AUROC is 0.777, against Jev's 0.784. Its calibration error falls to 0.025, which is lower than Jev's own.

Gold labels still give a more accurate model, at 74.9\%. Most of that gain comes from abstaining less often when a correct holding is present (7.2\% against 24.6\%), and the gold-trained model picks a wrong holding more often (38.1\%). Measured by accuracy, Jev's probabilities close about two-thirds of the gap between the untrained model and the gold-trained one, without any labeled data.

\section{Discussion}
\label{sec:discussion}

\subsection{Two Ways to Fail}

The two kinds of model fail in opposite directions. Most generative LLMs rarely abstain and often pick a wrong holding. Jev abstains often and picks a wrong holding least. We think this follows from how each is trained. Generative LLMs are tuned and graded on tasks where answering is rewarded and abstaining is not~\cite{kalai2025hallucinate}, and more reasoning does not reliably help them abstain~\cite{kirichenko2025abstentionbench}. Decision models are trained to return probabilities that match outcomes. That gives a user something to act on. Their probability of none separates items with and without a correct holding (AUROC 0.73 to 0.89).

We do not read this as a verdict for one kind of model. Jev's caution means it abstains on about one in four items that do have a correct holding, and a decision model's confidence is not always a reliable sign of missing knowledge~\cite{shankaranarayana2026selfknowledge}.

\subsection{Accuracy Is the Wrong Scoreboard for Legal Authority}

A wrong citation and an unnecessary review are not equal errors. The first can reach a filing and end in sanctions~\cite{noland2025,whiting2026}. The second costs a lawyer the time it takes to look something up. Accuracy counts them the same, which is how a model that guesses can outscore one that abstains. Our earlier work made the same point for precedent treatment, where some misclassifications matter far more than others~\cite{demir2025validate}.

How much worse a wrong citation is depends on the user, the matter, and the court. A pro se litigant drafting a first motion and a firm filing an appellate brief do not face the same stakes. We therefore report the full trade-off rather than one score. Choosing an operating point is a decision about acceptable risk, and it belongs to the lawyer or firm using the tool, not to its vendor. That choice can only be made well if evaluations of legal AI tools report, next to accuracy, how often a tool cites an authority when none applies and how often it declines when one does.

\subsection{A Cheap Check, Not a Substitute}

A decision model costs a small fraction of a generative LLM call, so it can be asked for a second opinion on every answer. Accepting a generative LLM's answer only when a decision model agrees halved wrong citations in our test while keeping nearly all correct answers. This agrees with the observation that hard verification items tend to defeat some models but not others~\cite{demir2026parcel}, so disagreement is informative.

None of this removes the lawyer's duty to read what they cite. Courts have been clear that every cited authority must be read and verified, whatever its source~\cite{noland2025}. What a combined system can do is point that reading at the answers most likely to be wrong.

Knowing the case law is also not the same as respecting what the passage supports. Shown only a case's citation, with none of the surrounding passage, the strongest models still picked its holding in about three of four items, well above the one-in-five chance rate (Appendix~\ref{app:contamination}). They have learned much of the case law itself. But legal authority depends on what a court actually held in the opinion at hand. In verification, prior knowledge can even override the provided text~\cite{demir2026parcel}. Our abstain condition rewards a model only for recognizing that the holding in front of it is not the one the court relied on.

\subsection{Running the Check Locally}

Many firms cannot send client material to external services, which rules out hosted models for much of their work~\cite{demir2025legalguardian}. The distillation result suggests one way forward. Trained on Jev's probabilities alone, with no labeled data, a local Qwen3-8B model reproduced Jev's accuracy and abstention behavior and became well calibrated, after 90 minutes of training on one workstation GPU. The teacher's probabilities can be collected on public material, such as published opinions, and the student can then run on the firm's own hardware. The student also inherits the teacher's mistakes, so a firm should test it on its own work before relying on it.

\section{Limitations}
\label{sec:limitations}

\textbf{Contamination.} CaseHOLD is built from public opinions decided before 2021, and the training data of the closed models we test is not disclosed. Our perturbation checks show that models know much of the case law but do not depend on recognizing the case; they cannot rule out that the models saw these passages in training.

\textbf{A constructed absence.} In our abstain condition, the correct holding is replaced by a real holding from another case. In practice, authority goes missing in other ways as well: a case may have been overruled, may come from the wrong jurisdiction, or may simply not exist. Our condition measures one controlled form of absence, not all of them.

\textbf{The cost of errors.} The values of $k$ we report are illustrative. We know of no empirical estimate of how much worse a wrong citation is than an unnecessary review, and the right value will differ across practice settings.

\textbf{Scope.} We test one task, United States case law in English, and one prompt per kind of model. Generative LLMs may behave differently under other prompts, reasoning settings, or retrieval, and the models we test will change; our results describe versions available in October 2026. Each distillation configuration was trained once, on 10,000 items, so its results carry more uncertainty than those of the hosted models.

\section{Conclusion}
\label{sec:conclusion}

We asked what generative LLMs and System One decision models do when the authority a court relied on is not among the candidates. Most generative LLMs still pick a holding, some in more than four of five such cases, while the most cautious decision model does so in fewer than one in three, at a fraction of the cost. Neither kind of model is reliable on its own, but pairing them halves the wrong citations of the strongest generative LLM we tested, and a decision model's probabilities can teach a local model to behave the same way without labeled data.

What calibrated probabilities offer is not correctness but a dial: the user can decide how cautious the system should be. Lincoln's advice to young lawyers was to be honest before anything else. For a research tool, honesty starts with saying when it cannot find the authority.

\bibliographystyle{ACM-Reference-Format}
\bibliography{refs}

\appendix
\section{Prompts}
\label{app:prompts}

All models receive the same instruction. Placeholders are in angle brackets; the option \texttt{NONE} appears only in the abstain condition.

\paragraph{Instruction.}
\begin{quote}\small
A court opinion cites a case and the parenthetical describing that case's holding has been replaced by \textless HOLDING\textgreater. Which holding statement is the one the citing court gave for that case?
\end{quote}

\paragraph{System One decision models.} One request per item to the decisions interface, with the passage as state and one choice question:
{\small
\begin{verbatim}
{"model": "<model id>",
 "state": {"citing_passage": "<citing_prompt>"},
 "questions": {"holding": {
   "type": "choice",
   "instructions": "<instruction>",
   "criteria": {"A": "<holding_0>", "B": "<holding_1>",
                "C": "<holding_2>", "D": "<holding_3>",
                "E": "<holding_4>",
                "NONE": "None of the listed holdings is the one
                         the cited case is being cited for in
                         this passage."}}}}
\end{verbatim}
}

\paragraph{Generative LLMs.} One user message per item, no system prompt:
{\small
\begin{verbatim}
<instruction>

Citing passage:
<citing_prompt>

Options:
A. <holding_0>
B. <holding_1>
C. <holding_2>
D. <holding_3>
E. <holding_4>
NONE. None of the listed holdings is the one the cited case
      is being cited for in this passage.

Reply with only the option key (A, B, C, D, E, NONE).
\end{verbatim}
}
The reply was constrained by a JSON schema with a single field, \texttt{answer}, whose value must be one of the option keys. Requests used temperature 0 and a maximum of 4,096 output tokens.

\section{Pilot Study and Additional Results}
\label{app:pilot}

\paragraph{Stated confidence.} Before the main study, we ran 1,000 randomly chosen test items with GPT-6 Luna prompted to return an answer together with its confidence from 0 to 100. In the abstain condition, it picked a holding in 60\% of the items without a correct one, and its stated confidence on those wrong picks averaged 0.91. Its expected calibration error was 0.29, against 0.07 and 0.08 for the two decision models run on the same items. Abstaining whenever its stated confidence fell below a high threshold (0.90) reduced its wrong picks to a level similar to a decision-model veto, but such a threshold has to be found on labeled data. Because generative LLMs do not state confidence in normal use, the main study asks them for an answer only.

\paragraph{Distillation into a smaller encoder.} We also fine-tuned the legal BERT model released with CaseHOLD~\cite{zheng2021casehold} on the full training split (42,509 items), three seeds per setting. Trained on gold labels, it reached 75.7\% on the original task and 61.3\% on the abstain condition. Adding Jev's probabilities to the training loss raised original-task accuracy by 0.7 points and halved its calibration error (0.043 to 0.022); on the abstain condition, it made the model abstain more often without improving accuracy. Because this encoder is already nearly as accurate as Jev on CaseHOLD, Jev has little to add. The larger gain for Qwen3-8B (Section~\ref{sec:results}) comes from a student that starts far below its teacher.

\section{Contamination Checks}
\label{app:contamination}

CaseHOLD is built from public opinions that predate the models we test, so a model may recognize a cited case rather than reason from the passage. For closed models, whose training data is not disclosed, contamination cannot be measured directly~\cite{sainz2023contamination}. We therefore use perturbation tests as a proxy. They cannot show whether a model saw the data, only whether its answers depend on recognizing the case.

\citet{arvin2025holdings} introduced such a test for CaseHOLD. A language model rewrote each passage so that every case name and citation was replaced by a plausible fictitious one; GPT-4o's macro F1 fell only from 0.744 to 0.728, and 88\% of its answers were unchanged. We repeat this anonymization with GPT-6 Luna as the rewriter on a random sample of 500 test items, and we check each rewrite automatically. The cited case's citation must be found where the holding was removed, and the rewritten passage must stay close to the original. We keep the 406 items that pass. Of the rest, 76 rewrites changed too much of the passage, and in 18 the citation could not be located. We add two variants on the same items: \emph{citation only}, in which the model sees only the cited case's citation, and \emph{citation removed}, in which that citation is replaced by \texttt{[citation omitted]}. High accuracy from the citation alone means a model knows the case; accuracy that survives anonymization and citation removal means it does not depend on that knowledge.

\begin{table}[t]
  \centering
  \small
  \caption{Accuracy (\%) on 406 test items in the original condition, with real citations, with case names and citations replaced by fictitious ones, with the cited case's citation removed, and with only that citation shown. Chance is 20\%.}
  \label{tab:contam}
  \begin{tabular}{lrrrr}
    \toprule
    Model & Original & Anonymized & Citation removed & Citation only \\
    \midrule
    Jev 1.14 & 78.8 & 77.1 & 78.3 & 45.3 \\
    pplx-decider v1.2 & 90.1 & 89.4 & 87.7 & 71.2 \\
    d1-ultra & 86.2 & 84.0 & 83.3 & 63.1 \\
    GPT-6 Luna Decisions & 73.8 & 73.3 & 72.3 & 41.7 \\
    Gemini 3.8 Flash & 88.7 & 85.5 & 83.5 & 74.4 \\
    Claude Haiku 5.5 & 78.1 & 76.1 & 75.6 & 50.0 \\
    GPT-6 Luna & 78.6 & 76.6 & 77.6 & 53.0 \\
    Mistral Small 2603 & 73.4 & 67.7 & 71.9 & 41.1 \\
    GLM 5.3 Flash & 77.6 & 72.7 & 74.4 & 46.6 \\
    GPT-4o-mini & 70.7 & 67.5 & 68.2 & 34.7 \\
    \bottomrule
  \end{tabular}
\end{table}

Models know much of the case law, but they do not depend on it (Table~\ref{tab:contam}). Shown only the citation of the cited case, the strongest models pick the right holding far above chance: 74.4\% for Gemini 3.8 Flash and 71.2\% for pplx-decider. Yet replacing every case name and citation with a fictitious one costs between 0.5 and 5.7 points, and removing the cited case's citation costs between 0.5 and 5.2. The decision models lose the least under anonymization (0.5 to 2.2 points). The strong results of pplx-decider are therefore not an artifact of recognizing the case, and its knowledge of case law is shared by the strongest generative LLM.

\section{Distillation Settings}
\label{app:distill}

We fine-tuned Qwen3-8B with LoRA (rank 16, $\alpha = 32$, dropout 0.05) on the attention and feed-forward projections, in bfloat16 with gradient checkpointing. Training used AdamW with a learning rate of $10^{-4}$, a cosine schedule with 5\% warmup, batch size 4 with 4 accumulation steps (effective batch 16), and one pass over 10,000 randomly sampled training items. The model saw the same instruction as the API models, with options lettered A--E, F for none, and thinking disabled. Passages were truncated to their last 600 tokens and each option to 160 tokens; one CaseHOLD holding exceeds 20,000 tokens. The Jev-trained model minimized the KL divergence from Jev's probabilities to the model's distribution over the option letters; the gold-trained model minimized cross-entropy against the correct option. Each run used one NVIDIA RTX A6000 (48\,GB).

\end{document}